\documentclass[runningheads]{llncs}
\usepackage{graphicx}
\pdfoutput=1

\usepackage{tikz}
\usepackage{comment}
\usepackage{amsmath,amssymb} 
\usepackage{color}

\usepackage[width=122mm,left=12mm,paperwidth=146mm,height=193mm,top=12mm,paperheight=217mm]{geometry}

\usepackage{mathtools}
\usepackage{multirow}
\usepackage{appendix}
\usepackage{xcolor}
\usepackage[normalem]{ulem}
\useunder{\uline}{\ul}{}

\def\mb#1{\mathbf{#1}}
\usepackage{multirow}
\def\fig#1{Fig.~\ref{fig:#1}}
\def\tab#1{Table~\ref{tab:#1}}
\def\sect#1{Sec.~\ref{sec:#1}}
\def\Eq#1{Eq.~(\ref{eq:#1})}

\def\mypar#1{\vspace{1mm}{\noindent\bf #1.}\hspace{1mm}}

\begin{document}

\pagestyle{headings}
\mainmatter
\def\ECCVSubNumber{7028}  

\title{Conditional-Flow NeRF: \\ Accurate 3D Modelling with Reliable Uncertainty Quantification} 

\titlerunning{Conditional-Flow NeRF}

\author{Jianxiong Shen\inst{1} \and Antonio Agudo\inst{1} \and Francesc Moreno-Noguer\inst{1} \and Adria Ruiz\inst{2}}
\authorrunning{J. Shen et al.}
\institute{Institut de Robòtica i Informàtica Industrial, CSIC-UPC,
Barcelona, Spain \email{\{jshen,aagudo,fmoreno\}@iri.upc.edu} \and Seedtag, Spain \\ \email{adriaruiz@seedtag.com}}

\maketitle

\begin{abstract}
A critical limitation of current methods based on Neural Radiance Fields (NeRF) is that they are unable to quantify the uncertainty associated with the learned appearance and geometry of the scene. This information is paramount in real applications such as medical diagnosis or autonomous driving where, to reduce potentially catastrophic failures, the confidence on the model outputs must be included into the decision-making process. In this context, we introduce Conditional-Flow NeRF (CF-NeRF), a novel probabilistic framework to incorporate uncertainty quantification into NeRF-based approaches. For this purpose, our method learns a distribution over all possible radiance fields modelling which is used to quantify the uncertainty associated with the modelled scene. In contrast to previous approaches enforcing strong constraints over the radiance field distribution, CF-NeRF learns it in a flexible and fully data-driven manner by coupling Latent Variable Modelling and Conditional Normalizing Flows. This strategy allows to obtain reliable uncertainty estimation while preserving model expressivity. Compared to previous state-of-the-art methods proposed for uncertainty quantification in NeRF, our experiments show that the proposed method achieves significantly lower prediction errors and more reliable uncertainty values for synthetic novel view and depth-map estimation.

\end{abstract}

\section{Introduction}
\label{sec:intro}

\begin{figure*}[t]
    \centering
    \includegraphics[width=0.9\textwidth]{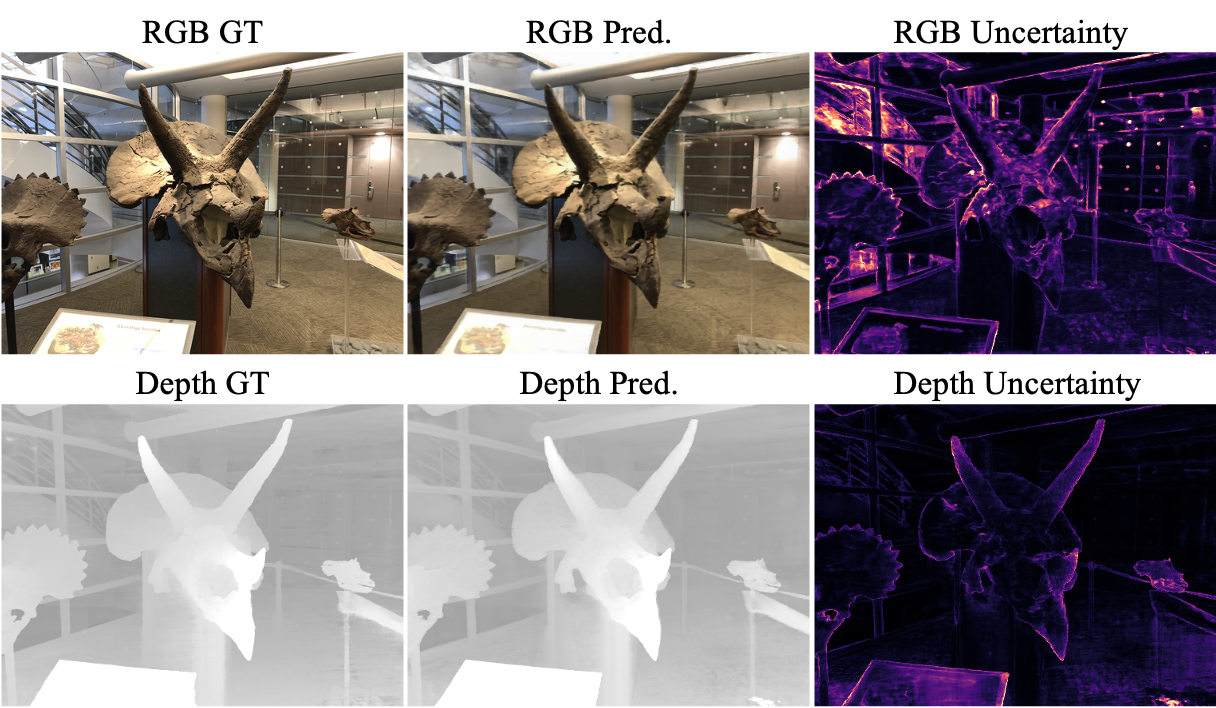}
    \vspace{-2mm}
    \caption{ Ground Truth (Left), Predictions (Center) and Uncertainty Maps (Right) obtained by our proposed CF-NeRF in novel view synthesis and depth-map estimation.
    }
    \label{fig:teaser}
\end{figure*}

Neural fields~\cite{neural_field} have recently gained a lot of attention given its ability to encode implicit representations of complex 3D scenes using deep neural networks. Additionally, they have been shown very effective in addressing multiple problems such as 3D reconstruction~\cite{OccupancyNetworks,deepsdf}, scene rendering~\cite{neural_volumes,nerfw} or human body representation~\cite{neural_fields_human_0,neural_fields_human_1}. Among different methods built upon this framework~\cite{neural_field_paper0,DeepRV,srns}, Neural Radiance Fields (NeRF)~\cite{nerf} has obtained impressive results in generating photo-realistic views of 3D scenes and solving downstream tasks such as depth estimation~\cite{nerfingmvs}, scene editing~\cite{scene_edit_1,scene_edit_2} or pose prediction~\cite{inerf,anerf}. 

Despite its increasing popularity, a critical limitation of NeRF  holding back its application to real-world problems has been recently pointed out by \cite{snerf}. Concretely, current NeRF-based methods are not able to quantify the uncertainty  associated with the model estimates. This information is crucial in several scenarios such as robotics \cite{robotics_0,robotics_1},  medical diagnosis~\cite{medical_diagnosis} or autonomous driving~\cite{autonomous_driving} where, to reduce potentially catastrophic failures, the confidence on the model outputs must be included into the decision-making process. 

To address this limitation, recent works~\cite{nerfw,snerf} have explored different strategies to incorporate uncertainty quantification into NeRF. Remarkably, Stochastic NeRF \cite{snerf} (S-NeRF) has recently obtained state-of-the-art results by employing a  probabilistic model to learn a simple distribution over radiance fields. However, in order to make this problem tractable, S-NeRF makes strong assumptions over the aforementioned distribution, thus limiting model expressivity and leading to sub-optimal results for complex 3D scenes. 

\begin{figure}[t!]
\centering
  \includegraphics[width=0.95\textwidth]{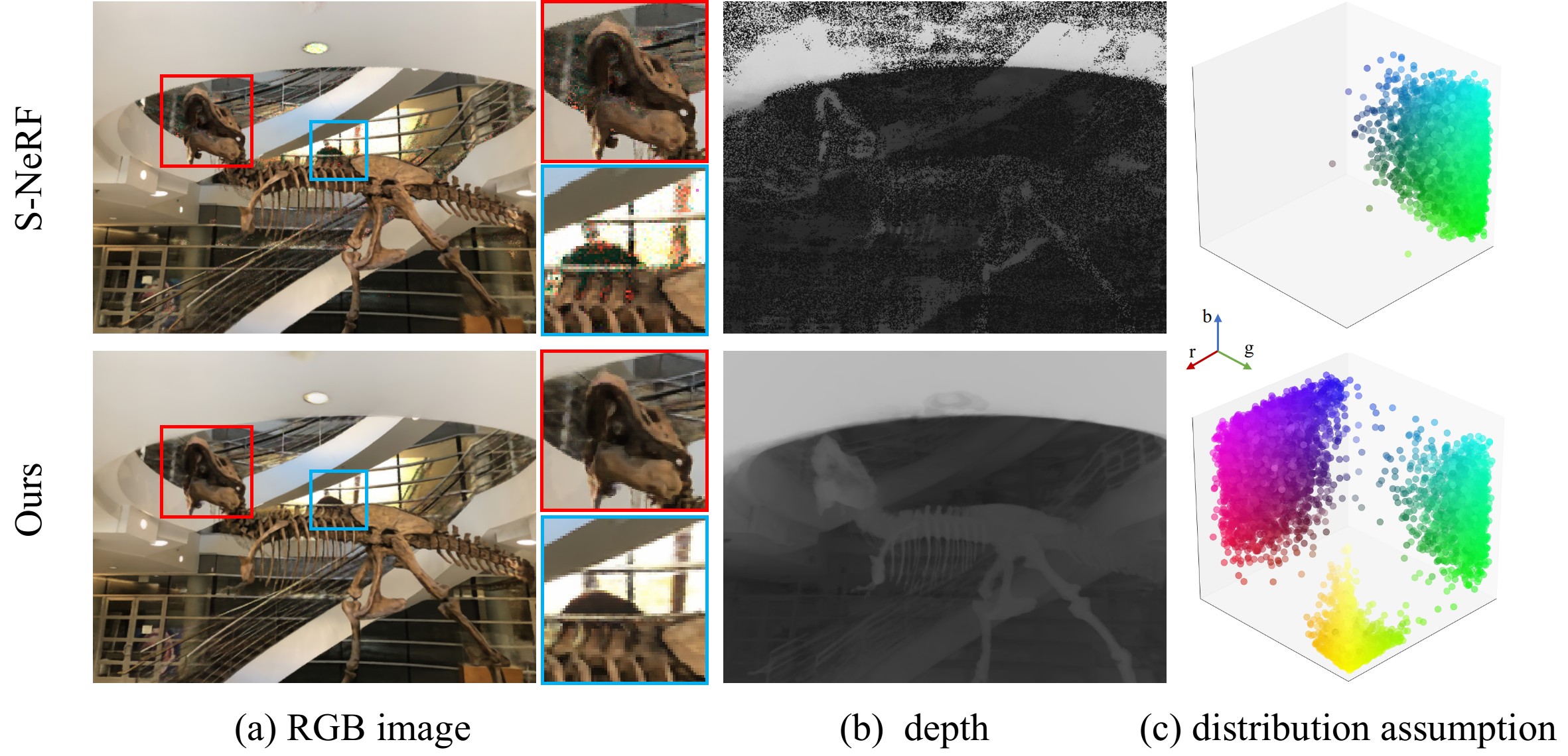}
  \vspace{-3mm}
  \caption{Comparison between our CF-NeRF and the previous state-of-the-art method S-NeRF~\cite{snerf} on (a) RGB images and (b) depth-maps. S-NeRF generates noisy results due to the strong assumptions made over the radiance field distribution. In contrast, CF-NeRF renders more realistic and smooth synthetic views. (c) Illustration of the  radiance distributions that can be modelled by S-NeRF and CF-NeRF. While the first is only able to represent distributions with a simple form, CF-NeRF can model arbitrary complex distributions by using Conditional Normalizing Flows.} 
  \label{fig:fig2}
\end{figure}

In this context, we propose Conditional-Flow NeRF (CF-NeRF), a novel probabilistic framework to incorporate uncertainty quantification without sacrificing model flexibility. As a result, our approach enables both effective 3D modelling and reliable uncertainty estimates (see \fig{teaser}). In the following, we summarise the main technical contributions of our method:

\vspace{1mm}
\noindent\textbf{Modelling Radiance-Density Distributions with Normalizing Flows:} In contrast to S-NeRF where the radiance and density in the scene are assumed to follow simple distributions (see \fig{fig2}(c-Top)), our  method learns them in a flexible and fully data-driven manner using conditional normalizing flows \cite{normalizing_flow_2}. This allows CF-NeRF to learn arbitrarily complex radiance and density distributions (see \fig{fig2}(c-Bottom)) which have the capacity to model scenes with complex geometry and appearance.

\vspace{1mm}
\noindent\textbf{Latent Variable Modelling for Radiance Fields Distributions:} Motivated by De Finetti's Theorem~\cite{De_Finetti}, CF-NeRF incorporates a global latent variable in order to efficiently model the joint distribution over the radiance-density variables for all the spatial locations in the scene. This contrasts with the approach followed by S-NeRF, where these variables are considered independent for different 3D coordinates. Intuitively, the independence assumption violates the fact that changes in the density and radiance between adjacent locations are usually smooth. As a consequence, S-NeRF tends to generate low-quality and noisy predictions (see \fig{fig2}(a,b)-Top). In contrast, the global latent variable used in CF-NeRF allows to efficiently model the complex joint distribution of these variables, leading to spatially-smooth uncertainty estimates and more realistic results (\fig{fig2}(a,b)-Bottom).

In our experiments, we evaluate CF-NeRF over different benchmark datasets containing scenes with increasing complexity. Our qualitative and quantitative results show that our method obtains significantly better uncertainty estimations for rendered views and predicted depth maps compared to S-NeRF and other previous methods for uncertainty quantification. Additionally, CF-NeRF provides better image quality and depth estimation precision. These results confirm that our method is able to incorporate uncertainty quantification into NeRF without sacrificing model expressivity.

\section{Related Work}
\label{sec:related_work}

\mypar{Neural Radiance Fields} NeRF~\cite{nerf} has become a popular approach for 3D scene modelling due to its simplicity and its impressive performance on several tasks. By using  a sparse collection of 2D views, NeRF learns a deep neural network encoding a representation of the 3D scene. This network is composed of a set of fully-connected layers that output the volume density and emitted radiance for any input 3D spatial location and 2D viewing direction in the scene. Subsequently, novel views are generated by applying volumetric rendering ~\cite{volume_rendering} to the estimated density and radiance values obtained by the network.

Recently, several extensions over original NeRF have been proposed  ~\cite{nerf++,Animatable_nerf,donerf,VAE_nerf,Transparent_nerf,Implicit_nerf,Fourier_nerf,GRAFGR,pixel_nerf}. For instance, different works have focused on accelerating NeRF training ~\cite{faster_nerf_1,faster_nerf_2,faster_nerf_3} or modelling dynamic scenes ~\cite{dyn_nerf_1,dyn_nerf_2,dyn_nerf_3,dyn_nerf_4,dyn_nerf_5}.  Despite the advances achieved at different levels, the application of current NeRF-based methods in real scenarios is still limited since they are unable to quantify the uncertainty associated with the rendered views or estimated geometry. Our proposed CF-NeRF explicitly addresses this problem and can be easily combined with most of current NeRF-based approaches. 

\vspace{1mm}
\mypar{Uncertainty Estimation in Deep Learning}
Uncertainty estimation has been extensively studied in deep learning~\cite{unc_in_dl_1,unc_in_dl_2,unc_in_dl_3,unc_in_dl_4,unc_in_dl_5} and have been applied to different computer vision tasks ~\cite{ause1,ause2,unc_in_cv_1}. Early works proposed to use \textit{Bayesian Neural Networks} (BNN)~\cite{bnn_1,bnn_2} to estimate the uncertainty of both network weights and outputs by approximating their marginal distributions.  However, training BNNs is typically difficult and computationally expensive. As a consequence, their use in large deep neural networks is limited.

More recently, efficient strategies have been proposed to incorporate uncertainty estimation into deep neural networks ~\cite{Laplace,Backprop}. Among them, MC-Dropout ~\cite{mc_dropout} and Deep Ensembles ~\cite{deep_ensemble} are two of the most popular approaches given that they are agnostic to the specific network architecture  ~\cite{dropout_1,dropout_2,ensemble_1,ensemble_2}. More concretely, MC-Dropout adds stochastic dropout during inference into the intermediate network layers. By doing multiple forward passes over the same input with different dropout configurations, the variance over the obtained set of outputs is used as the output uncertainty. On the other hand, Deep Ensembles applies a similar strategy by computing a set of outputs from multiple  neural networks that are independently trained using different parameters initializations. In contrast to  MC-Dropout and Deep Ensembles, however, we do not train independent networks to generate the samples. Instead, CF-NeRF explicitly encodes the distribution of the radiance fields into a single probabilistic model.

\mypar{Uncertainty Estimation in NeRF}
Recently, some works~\cite{nerfw,snerf} have attempted to incorporate uncertainty estimation into Neural Radiance Fields. NeRF-in-the-Wild (NeRF-W)~\cite{nerfw} modelled uncertainty at pixel-level in order to detect potential transient objects such as cars or pedestrians in modelled scenes.  For this purpose, NeRF-W employs a neural network to estimate an uncertainty value for each 3D spatial location. Subsequently, it computes a confidence score for each pixel in a rendered image by using an alpha-compositing strategy similar to the one used to estimate the pixel color in original NeRF. However, this approach is not theoretically grounded since there is no intuitive justification to apply the volume rendering process over uncertainty values. Additionally, NeRF-W is not able to evaluate the confidence associated with estimated depth maps. 

On the other hand, S-NeRF~\cite{snerf} has achieved state-of-the-art performance in quantifying the uncertainty of novel rendered views and depth-maps. Instead of learning a single radiance field as in original NeRF~\cite{nerf}, S-NeRF models a distribution  over all the possible radiance fields explaining the scene. During inference, this distribution is used to sample multiple color or depth predictions and compute a confidence score based on their associated variance. However, S-NeRF imposes strong constraints over the radiance field distribution and, as a consequence, this limits its ability to model scenes with complex appearance and geometry. In contrast, the proposed CF-NeRF combines a conditional normalizing flow and latent variable modelling to learn arbitrary complex radiance fields distributions without any prior assumption. 

\mypar{Complex Distribution Modelling with Normalizing Flows}
Normalizing Flows (NF) have been extensively used to model complex distributions with unknown explicit forms. For this purpose, NF uses a sequence of invertible functions to transform samples from a simple known distribution to variables following an arbitrary probability density function. Additionally, the change-of-variables formula can be used to compute the likelihood of different samples according to the transformed distribution. Given its flexibility, Normalizing Flows have been used in several 3D modelling tasks. For instance, \cite{PointFlow,CFlow} introduced different flow-based generative models to model 3D point cloud distributions and sample from them. More recently, \cite{RegNeRFRN} used a normalizing flow in order to avoid color shifts at unseen viewpoints in the context of NeRF. To the best of our knowledge, our proposed CF-NeRF is the first approach employing Normalizing Flows in order to learn radiance fields distributions for 3D scene modelling. 

\section{Deterministic and Stochastic Neural Radiance Fields}
\label{sec:nerf}

\mypar{Deterministic Neural Radiance Fields} Standard NeRF \cite{nerf} represents a 3D volumetric scene as $\mathcal{F} = \{ (\mathbf{r}(\mathbf{x},\mathbf{d}), \alpha(\mathbf{x})): \mathbf{x} \in \mathbb{R}^3, \mathbf{d} \in \mathbb{R}^2$, 
where $\mathcal{F}$ is a set containing the volume density $\alpha(\mathbf{x}) \in \mathbb{R}^+$  and RGB radiance $\mathbf{r}(\mathbf{x},\mathbf{d}) \in \mathbb{R}^3$ from all the spatial locations $\textbf{x}$  and view directions $\mathbf{d}$ in the scene.

In order to implicitly model the infinite set $\mathcal{F}$, NeRF employs a neural network $f_{\boldsymbol{\theta}}(\mathbf{x},\mathbf{d})$ with parameters ${\boldsymbol{\theta}}$ which outputs the density $\alpha$ and radiance $\mathbf{r}$ for any given input location-view pair $\{\mathbf{x},\mathbf{d}\}$. Using this network, NeRF is able to estimate the color $\mb c(\mathbf{x}_o,\mathbf{d})$ for any given pixel defined by a 3D camera position $\mathbf{x}_o$ and view direction $\mathbf{d}$ using the volumetric rendering function:
\begin{equation}
\label{eq:vol_rendering}
\mb c (\mathbf{x}_o,\mathbf{d}) = \int_{t_n}^{t_f} T(t) \alpha(\mathbf{x}_t) \mathbf{r}(\mathbf{x}_t, \mathbf{d}) dt, \hspace{2mm} \text{where} \hspace{2mm} T(t) = \text{exp}(-\int_{t_n}^{t} \alpha(\mathbf{x}_s) ds),
\end{equation}
where $\mathbf{x}_t = \mathbf{x}_o + t \mb d$ corresponds to 3D locations along a ray with direction $\mb d$ originated at the camera origin and intersecting with the pixel at $\mathbf{x}_o$.

During training, NeRF optimizes the network parameters ${\boldsymbol{\theta}}$ using \Eq{vol_rendering} by leveraging Maximum Likelihood Estimation (MLE) over a training set $\mathcal{T}$ containing ground-truth images depicting views of the scene captured from different camera positions. More details about NeRF and its learning procedure can be found in the original paper~\cite{nerf}.

\vspace{1mm}
\mypar{Stochastic Neural Radiance Fields}  Deterministic NeRF is not able to provide information about the underlying uncertainty associated with the rendered views or estimated depth maps. The reason is that the network $f_{\boldsymbol{\theta}}(\mathbf{x},\mathbf{d})$ is trained using Maximum Likelihood Estimation and thus, the learning process performs a single point estimate over all the plausible radiance fields  $\mathcal{F}$ given the training set $\mathcal{T}$. As a consequence, model predictions are deterministic and it is not possible to sample multiple outputs to compute their associated variance.

To address this limitation, S-NeRF \cite{snerf} employs Bayesian Learning to model the posterior distribution  $p_{\boldsymbol{\theta}}(\mathcal{F} | \mathcal{T})$ over all the plausible radiance fields explaining the scene given the training set. In this manner, uncertainty estimations can be obtained during inference by computing the variance over multiple predictions obtained from different radiance fields $\mathcal{F}$ sampled from this distribution.

In order to make this problem tractable, S-NeRF uses Variational Inference to approximate the posterior $p_{\boldsymbol{\theta}}(\mathcal{F} | \mathcal{T})$ with a parametric distribution $q_{\boldsymbol{\theta}}(\mathcal{F})$ implemented by a deep neural network. However, S-NeRF imposes two limiting constraints over $q_{\boldsymbol{\theta}}(\mathcal{F})$. Firstly, it models the radiance $\mb {r}$ and density $\alpha$ for each location-view pair in the scene as independent variables. In particular, the probability of a radiance field given $\mathcal{T}$ is modelled with a fully-factorized distribution: 
\begin{equation}
    \label{eq:apost_definition}
    q_{\boldsymbol{\theta}}(\mathcal{F}) = \prod_{\mathbf{x} \in \mathbb{R}^3} \prod_{\mathbf{d} \in \mathbb{R}^2} q_{\boldsymbol{\theta}}(\mathbf{r}|\mathbf{x},\mathbf{d}) q_{\boldsymbol{\theta}}(\alpha|\mathbf{x}).
\end{equation}
Whereas this conditional-independence assumption simplifies the optimization process, it ignores the fact that radiance and density values in adjacent spatial locations are usually correlated. As a consequence, S-NeRF tends to render noisy and low-quality images and depth maps (see \fig{fig2}).

\section{Conditional Flow-based Neural Radiance Fields}
\label{sec:f_nerf}

Our proposed Conditional Flow-based Neural Radiance Fields incorporates uncertainty estimation by using a similar strategy than S-NeRF. In particular, CF-NeRF learns a parametric distribution $q_{\boldsymbol{\theta}}(\mathcal{F})$ approximating the posterior distribution $p(\mathcal{F}|\mathcal{T})$ over all the possible radiance fields given the training views. Different from S-NeRF, however, our method does not assume a simple and fully-factorized distribution for $q_{\boldsymbol{\theta}}(\mathcal{F})$, but it relies on Conditional Normalizing Flows and Latent Variable Modelling to preserve model expressivity. In the following, we discuss the technical details of our approach (\sect{cf_nerf}), the optimizing process following S-NeRF (\sect{optimization}) and inference procedure used to compute the uncertainty associated with novel rendered views and depth maps (\sect{inference}).

\subsection{Modelling Flexible Radiance Field Distributions with CF-NeRF}
\label{sec:cf_nerf}
\mypar{Radiance Field distribution with Global Latent Variable}
\label{sec:global_latent_variable}
As discussed in \sect{nerf}, S-NeRF formulation assumes that radiance and density variables at each spatial location in the field follow a fully-factorized distribution ( \Eq{apost_definition}). As a consequence, they are forced to be independent from each other. This simplification was motivated by the fact that modelling the joint distribution over the infinite set of radiance and density variables in the field $\mathcal{F}$ is not trivial. To address this problem, our proposed CF-NeRF leverages De Finetti's representation theorem \cite{De_Finetti}. In particular, the theorem states that any joint distribution over sets of exchangeable variables can be written as a fully-factorized distribution conditioned on a global latent variable. Based on this observation, we define the approximate posterior $q_{\boldsymbol{\theta}}(\mathcal{F})$: 
\begin{equation}
\label{eq:factored_distribution}
    q_{\boldsymbol{\theta}}(\mathcal{F}) = \int_{\mb z} q_\vartheta( \mb z) q_{\boldsymbol{\theta}}(\mathcal{F}| \mb z)d \mb z  
    = \int_{\mb z} q_\vartheta(\textbf{z}) \prod_{\mathbf{x} \in \mathbb{R}^3} \prod_{\mathbf{d} \in \mathbb{R}^2} q_{\boldsymbol{\theta}}(\mathbf{r}|\mathbf{x},\mathbf{d},\mb z) q_{\boldsymbol{\theta}}(\alpha|\mathbf{x}, \mb z)d \mb z \, ,
\end{equation}
where the latent variable $\mb z$ is sampled from a Gaussian prior $q_\vartheta(\mb z)=\{\mb \mu_z , \mb \sigma_z\}$ with learned mean and variance. In this manner, all the 3D spatial-locations $\mathbf{x}$ and viewing-directions $\mathbf{d}$ generated from the same shared latent variable $\mb z$ and thus, they are conditionally dependent between them. This approach allows CF-NeRF to efficiently model the distribution over the radiance-density variables in the radiance field without any independence assumption. It is also worth mentioning that De Fineti's theorem has been also used in previous works learning joint distributions over sets for 3D point cloud modelling \cite{DiscretePF}.

\mypar{Radiance-Density Conditional Flows}
\label{sec:conditional_flow}
Given that the radiance-density distributions can be highly complicated for complex scenes, modelling them with variants of Gaussian distributions as in Stochastic NeRF can lead to sub-optimal results. For this reason, CF-NeRF models $q_{\boldsymbol{\theta}}(\mathbf{r}|\mathbf{x},\mathbf{d},\mb z)$ and $q_{\boldsymbol{\theta}}(\alpha|\mathbf{x},\mb z)$ using a Conditional Normalizing Flow (CNF) \cite{sylvester}. This allows to learn arbitrarily complex distributions in a fully data-driven manner without any prior assumption. 

\begin{figure*}[t]
    \centering
    \includegraphics[width=1.0\textwidth]{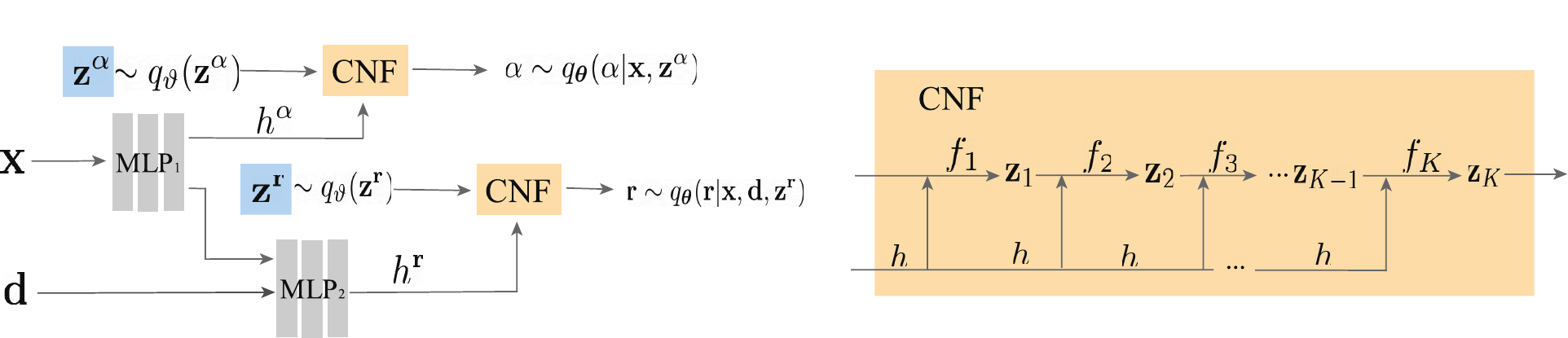}
    \vspace{-5mm}
    \caption{ (Left) We use two Conditional Normalizing Flow (CNF) to sample radiance and density values from distributions $q_{\boldsymbol{\theta}}(\mathbf{r}|\mathbf{x},\mathbf{d},\mb z)$ and $q_{\boldsymbol{\theta}}(\alpha|\mathbf{x},\mb z)$, respectively. Each CNF computes a transformation of a sample from the latent distribution $q_\psi(\mathbf{z})$ conditioned to an embedding $\mb h$. This embedding which is computed by an MLP with the location-view pair $(\mb x, \mb d)$ as input. (Right) Each CNF is composed by a sequence of invertible transformation functions $f_{1:K}$ conditioned on $\mb h$.}
    \label{fig:conditional_nf}
\end{figure*}

For any input location-direction pair $(\mathbf{x},\mathbf{d})$, we use a sequence of $K$ CNFs that transform the global latent variable $\mb z$ into samples from the radiance $\textbf{r}$ or density $\alpha$ distribution $q_{\boldsymbol{\theta}}(\mathbf{r}|\mathbf{x},\mathbf{d},\mb z)$ or $q_{\boldsymbol{\theta}}(\alpha|\mathbf{x},\mb z)$. More formally, each flow is defined as an invertible parametric function $\mathbf{z}_k = f_k(\mathbf{z}_{k-1},\mathbf{\mb x, \mb d})$, where $f_k$ maps a random variable $\textbf{z}_{k-1}$ into another one $\textbf{z}_k$ with a more complex distribution. Note that the each flow $f_k$ is conditioned on the location and view direction $(\mathbf{x},\mathbf{d})$. Finally, $\textbf{z}_K$ is followed by a Sigmoid and Softplus activation functions for radiance and density samples, respectively. This process is detailed in \fig{conditional_nf}.

Using the introduced CNF, radiance and density probabilities $q_{\boldsymbol{\theta}}(\mathbf{r}|\mathbf{x},\mathbf{d},\mb z)$ and $q_{\boldsymbol{\theta}}(\alpha|\mathbf{x},\mb z)$ can be computed with the change-of-variables formula typically used in normalizing flows as:
\begin{align}
    \label{eq:radiance_probability}
    q_{\boldsymbol{\theta}}(\mb r|\mathbf{x},\mb d, \mb z) &= q_\vartheta(\mathbf{z}) \left| \det \frac{\partial \mathbf{z}}{\partial \mathbf{r}} \right|
    = q_\vartheta(\mathbf{z}) \left| \det \frac{\partial \mathbf{r}}{\partial \mathbf{z}_{K}} \right|^{-1} \prod_{k=1}^K \left| \det \frac{\partial \mathbf{z}_k}{\partial \mathbf{z}_{k-1}} \right|^{-1} ,
\end{align}
where $\left| \det (\partial \mathbf{z}_k / \partial \mathbf{z}_{k-1}) \right|$ measures the Jacobian determinant of the transformation function $f_k$. Note that  $q_{\boldsymbol{\theta}}(\alpha|\mathbf{x},\mb z)$ can be computed in a similar manner.


\begin{figure*}[t]
    \centering
    \includegraphics[width=1.0\textwidth]{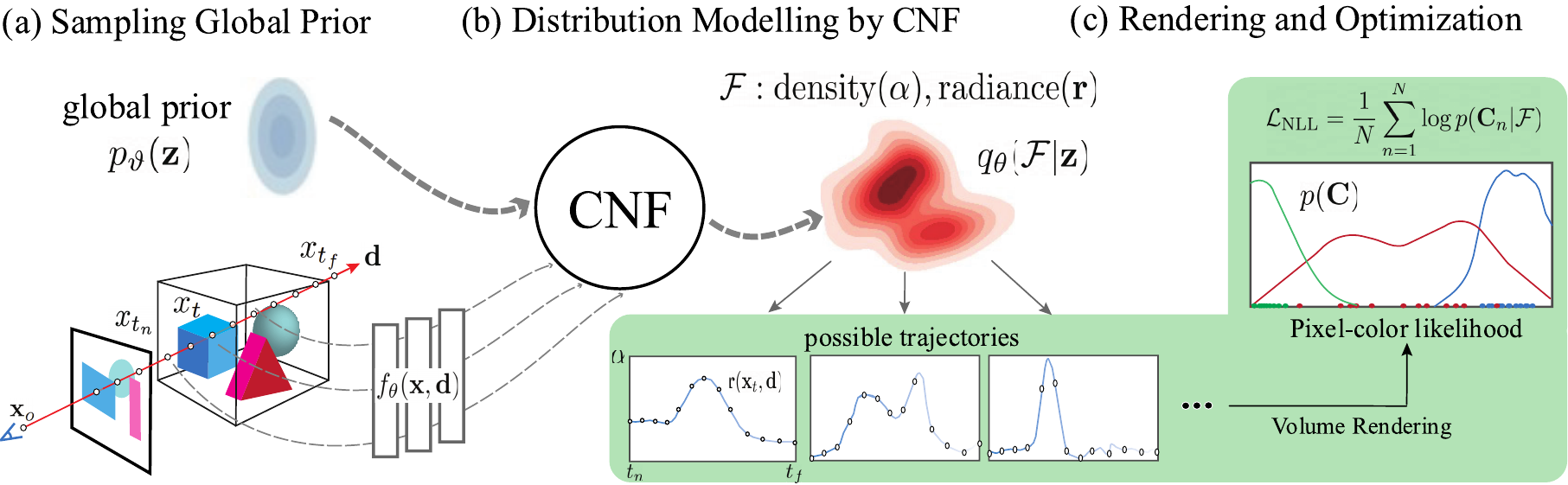}
    \vspace{-5mm}
    \caption{\textbf{Illustration of our pipeline for the inference and computation of the log-likelihood on the pixel color.} (a) We sample a set of variables $\textbf{z}$ from the global latent distribution. (b) Given each spatial location $\mathbf{x}$ along a camera ray with viewing direction $\mathbf{d}$, we can generate a set of density and radiance values by passing each $\textbf{z}$ variable through our proposed CNF. (c) These values can be represented as a set of different density-radiance trajectories along the ray corresponding to each $\textbf{z}$, followed by volume rendering techniques to composite each trajectory into a RGB value. Finally, these RGB values are used to compute the log-likelihood for the pixel color and also estimate the model prediction and its associated uncertainty using their mean and variance during inference.}
    \label{fig:pipeline}
\end{figure*}

\subsection{Optimizing Conditional-Flow NeRF}
\label{sec:optimization}

We adopt a Variational Bayesian approach to learn the parameters of the posterior distribution $q_{\boldsymbol{\theta}}(\mathcal{F})$ defined in \Eq{factored_distribution}. Note that the optimized $\mathbf{\theta}$ corresponds to the parameters of the CNFs defined in the previous section. More formally, we solve an optimization problem where we minimize the Kullback-Leibler (KL) divergence between $q_{\boldsymbol{\theta}}(\mathcal{F})$ and the true posterior distribution $p(\mathcal{F} | \mathcal{T})$ of radiance fields given the training set: 
\begin{align}
    \label{eq:kl_fnerf}
    &\min_{{\boldsymbol{\theta}}} \hspace{1mm} \mathbb{KL}(q_{\boldsymbol{\theta}}(\mathcal{F}) || p( \mathcal{F} | \mathcal{T})) \nonumber \\
    &= \min_{{\boldsymbol{\theta}}} - \mathbb{E}_{q_{\boldsymbol{\theta}}(\mathcal{F})} \log p( \mathcal{T} | \mathcal{F}) +
    \mathbb{E}_{q_{\boldsymbol{\theta}}(\mathcal{F})} \log q_{\boldsymbol{\theta}}(\mathcal{F})   - \mathbb{E}_{q_{\boldsymbol{\theta}}(\mathcal{F})} \log p(\mathcal{F}) . 
\end{align}

The first term in \Eq{kl_fnerf} measures the expected log-likelihood of the training set $\mathcal{T}$ over the distribution of radiance fields $q_{\boldsymbol{\theta}}(\mathcal{F})$. The second term indicates the negative Entropy of the approximated posterior. Intuitively, maximizing the Entropy term allows uncertainty estimation by preventing the optimized distribution to degenerate into a deterministic function where all the probability is assigned into a single radiance field $\mathcal{F}$. Finally, the third term corresponds to the cross-entropy between $q_{\boldsymbol{\theta}}(\mathcal{F})$ and a prior over radiance fields $p(\mathcal F)$. Given that it is hard to choose a proper prior in this case, we assume $p(\mathcal F)$ to follow uniform distribution and thus, this term can be ignored during optimization. In the following, we detail how the first two terms are computed during training.

\mypar{Computing the log-likelihood}
\label{sec:log_likelihood} 
Assuming that the training set $\mathcal{T}$ is composed by $N$ triplets $\{\mb C_n,\mb x_n^o ,\mathbf{d}_n\}$ representing the color, camera origin and view direction of a given pixel in a training image, the log-likelihood term in \Eq{kl_fnerf} is equivalent to: $ \frac{1}{N} \sum_{n=1}^N \log p(\textbf{C}_n | \mathcal{F})$. In order to 
compute an individual $p(\mb C_n | \mathcal{F})$ we use the following procedure. 

Firstly, we sample a set of variables $\textbf{z}^{1:K}$ from the global latent distribution $q_\vartheta(\mb z)$. Secondly, given a training triplet $\{\mb C,\mb x^o,\mathbf{d}\}$ and a set of 3D spatial locations along a ray defined by $\textbf{x}_t=\textbf{x}_o + t\textbf{d}$, we sample a set of density  and radiance values $\{\alpha^{1:K}\}_{t_n:t_f}$ and $\{\textbf{r}^{1:K}\}_{t_n:t_f}$  by using the Conditional Flow introduced in \sect{cf_nerf}. In particular, we perform a forward pass over the flow with inputs $\textbf{z}^k$ conditioned to $\textbf{x}^*_t$ for all $t$. Subsequently, a set of $K$ color estimates $\hat{\textbf{C}}_{1:K}$ are obtained by using, for every $k$, the volume rendering formula in \Eq{vol_rendering} over the sampled radiance and density values  $\{\alpha^k\}_{t_n:t_f}$ and $\{\textbf{r}^k\}_{t_n:t_f}$, respectively. Finally, with the set of obtained color estimations $\hat{\textbf{C}}_{1:K}$, we use a non-parametric kernel density estimator as in \cite{snerf} to approximate the log-likelihood $\log p(\mb C | \mathcal{F})$ of the pixel color. An illustration of this procedure is shown in \fig{pipeline}. 

\mypar{Computing the Entropy term}
\label{sec:entropy}
In order to compute the entropy term in \Eq{kl_fnerf}, we use a Monte Carlo approximation with $M$ samples as:
{\footnotesize
\begin{align}
    \mathbb{E}_{q_{\boldsymbol{\theta}}(\mathcal{F})} \log q_{\boldsymbol{\theta}}(\mathcal{F}) &= \mathbb{E}_{q_{\boldsymbol{\vartheta}}(\mathbf{z})} \prod_{\mathbf{x} \in \mathbb{R}^3} \prod_{\mathbf{d} \in \mathbb{R}^2} \log q_{\boldsymbol{\theta}}(\mathbf{r}|\mathbf{x},\mathbf{d},\textbf{z}) +
    \mathbb{E}_{q_{\boldsymbol{\vartheta}}(\mathbf{z^{\alpha}})} \prod_{\mathbf{x} \in \mathbb{R}^3} \prod_{\mathbf{d} \in \mathbb{R}^2} \log q_{\boldsymbol{\theta}}(\alpha|\mathbf{x},\textbf{z}) \nonumber \\
    &\sim \frac{1}{M} \sum_{m=1}^M \bigg( \log q_{\boldsymbol{\theta}}(\mathbf{r}_m|\mathbf{x}_m,\mathbf{d}_m,\textbf{z}_m) + \log q_{\boldsymbol{\theta}}(\alpha_m|\mathbf{x}_m,\mb z_m) \bigg)
\end{align}}
where $\mb {z}_m$ are obtained from the latent variable distribution $q_{\boldsymbol{\vartheta}}(\mathbf{z})$ and $\{\mb {x}_m,\mb {d}_m\}$ are possible 3D locations and view directions in the scene which are randomly sampled. Finally, $\alpha_m$ and $\mb r_m$ are density and radiance values obtained by applying our conditional flow with inputs $(\mathbf{z}_m$,$\textbf{x}_m$,$\mb {d}_m)$. Their probabilities $q_{\boldsymbol{\theta}}(\alpha_m|\mathbf{x}_m,\mb z_m)$ and $q_{\boldsymbol{\theta}}(\mathbf{r}_m|\mathbf{x}_m,\mathbf{d}_m,\textbf{z}_m)$  can be computed by using \Eq{radiance_probability}.

\subsection{Inference and Uncertainty Estimation}
\label{sec:inference}

CF-NeRF allows us to quantify the uncertainty associated with rendered images and depth-maps. As described in \sect{log_likelihood}, given any camera pose, we can obtain a set of color estimates $\hat{\textbf{C}}_{1:K}$ for every pixel in a rendered image. These estimates are obtained by sampling a set of random variables $\textbf{z}_{1:K}$ from latent distribution $q_\vartheta(\mb z)$ and applying CF-NeRF. Finally, the mean and variance of the pixel colors over the K samples are treated as the predicted color and its associated uncertainty, respectively. For depth-map generation, we sample a sequence of density values $\{\alpha_{k}\}_{t_n:t_f}$ along a camera ray and we compute the expected termination depth of the ray as in ~\cite{nerf}. In this way, we obtain a set of depth-values $d_{1:K}$ for each pixel in the depth-map for the rendered image. As in the case of the RGB color, its mean and variance correspond to the estimated depth and its associated uncertainty.

\section{Experiments}

\mypar{Datasets} 
We conduct a set of exhaustive experiments over two benchmark databases typically used to evaluate NeRF-based approaches: the \textbf{LLFF} dataset from the original NeRF paper~\cite{nerf} and the Light Field \textbf{LF} dataset ~\cite{nerf++,lfdataset}. The first is composed of 8 relatively simple scenes with multiple-view forward-facing images. On the other hand, from the LF dataset we use 4 scenes: \textit{Africa, Basket, Statue} and  \textit{Torch}. The evaluation over the {LF} scenes is motivated by the fact that they have a longer depth range compared to the ones in  {LLFF} and, typically, they present more complicated geometries and appearance. As a consequence, the evaluation over LLFF gives more insights into the ability of NeRF-based methods to model complex 3D scenes.  Same as ~\cite{snerf}, we use a sparse number of scene views ($\sim$4) for training and the last adjacent views for testing. As discussed in the cited paper, training in an extremely low-data regime is a challenging task and provides an ideal setup to evaluate the ability of the compared methods to quantify the uncertainty associated with the model predictions. 
While the ground truth of depth is not available in the original datasets, we compute pseudo ground truth by using the  method in ~\cite{nerfingmvs} trained on all the available views per scene.

\mypar{Baselines} We compare our method with the state-of-the-art S-NeRF~\cite{snerf}, NeRF-W~\cite{nerfw} and two other methods, Deep-Ensembles(D.E.)~\cite{deep_ensemble} and MC-Dropout(Drop.)~\cite{mc_dropout}, which are generic approaches for uncertainty quantification in deep learning. For NeRF-W, we remove the latent embedding component of their approach and keep only the uncertainty estimation layers. For Deep-Ensembles, we train 3 different NeRF models in the ensemble in order to have a similar computational cost during training and testing compared to the rest of compared methods. In the case of  MC-Dropout, we manually add one dropout layer after each odd layer in the network to sample multiple outputs  using random dropout configurations. The described setup and hyper-parameters for the different baselines is the same previously employed in S-NeRF~\cite{snerf}.

\mypar{Implementations details}
\label{sec:implementations}
In order to implement CF-NeRF and the different baselines, we inherit the same network architecture and hyper-parameters than the one employed in the original NeRF\footnote{https://github.com/bmild/nerf}  implementation. In CF-NeRF, we use this architecture to compute the conditional part given $\mb x$ and $\mb d$.
Since CF-NeRF uses an additional normalizing flow following this network, we add additional layers for the rest of baselines so that they have a similar computational complexity than our method. During training and inference in CF-NeRF,  we sample 32 radiance-density pairs for mean and variance estimation for each ray. We optimize all the models for 100,000-200,000 steps with a batch size of 512 and uniformly sampled 128 points across each ray using Adam optimizer with default hyper-parameters.
See the supplementary material for additional details on hyper-parameters of our conditional normalizing flow. 

\mypar{Metrics} In our experiments, we address two different tasks using the compared methods: novel view synthesis and depth-map estimation. In particular, we evaluate the quality of generated images/depth-maps and the reliability of the quantified uncertainty using the following metrics:   

\textit{Quality metrics:} We use standard metrics in the literature. Specifically, we report PSNR, SSIM~\cite{ssim}, and LPIPS~\cite{lpips} for generated synthetic views. On the other hand, we compute RMSE, MAE and $\delta$-threshold~\cite{delta_threshold} to evaluate the quality of the generated depth-maps.

\textit{Uncertainty Quantification:} To evaluate the reliability of the uncertainty estimated by each model we report two widely used metrics for this purpose. Firstly, we use the negative log likelihood (NLL) which is a proper scoring rule ~\cite{nll1,nll2} and has been previously used to evaluate the quality of model uncertainty \cite{snerf}. 
Secondly, we evaluate the compared methods using sparsification curves~\cite{ause1,ause2,ause3}. Concretely, given an error metric (e.g. RMSE), we obtain two lists by sorting the values of all the pixels according to their uncertainty and the error computed from the ground-truth. By removing the top $t\% (t=1\sim 100)$ of the errors in each vector and repeatedly computing the average of the last subset, we can obtain the sparsification curve and the oracle curve respectively. The area between them is the AUSE, which evaluates how much the uncertainty is correlated with the predicted error. 

\subsection{Results over LLFF dataset}

\begin{table}[t]
\centering
\resizebox{0.8\textwidth}{!}{
\begin{tabular}{llcccccccc}
\hline
 &  & \multicolumn{3}{c}{quality metrics} &  & \multicolumn{3}{c}{uncertainty metrics} &  \\ \cline{3-5} \cline{7-9}
 &  & PSNR$\uparrow$ & SSIM$\uparrow$ & LPIPS$\downarrow$ &  & AUSE RMSE$\downarrow$ & AUSE MAE$\downarrow$ & NLL$\downarrow$ &  \\ \hline
D.E.~\cite{deep_ensemble} &  & \textbf{22.32} & {\ul 0.788} & 0.236 &  & 0.0254 & 0.0122 & 1.98 &  \\
Drop.~\cite{mc_dropout} &  & 21.90 & 0.758 & 0.248 &  & 0.0316 & 0.0162 & 1.69 &  \\
NeRF-W~\cite{nerfw} &  & 20.19 & 0.706 & 0.291 &  & 0.0268 & 0.0113 & 2.31 &  \\
S-NeRF~\cite{snerf} &  & 20.27 & 0.738 & {\ul 0.229} &  & { {\ul 0.0248}} & { {\ul 0.0101}} & { {\ul 0.95}} &  \\
CF-NeRF &  & {\ul 21.96} & \textbf{0.790} & \textbf{0.201} &  & \textbf{0.0177} & \textbf{0.0078} & \textbf{0.57} & \\ \hline
\end{tabular}}
\caption{Quality and uncertainty quantification metrics on rendered images over the LLFF dataset. Best results are shown in bold with second best underlined. See text for more details.}
\label{tab:table_llff_rgb_uncertainty}
\end{table}

Following a similar experimental setup than~\cite{snerf}, we firstly evaluate the performance of the compared methods on the simple scenes in the LLFF dataset. 
All the views in each scene are captured in a restricted setting where all cameras are forward-facing towards a plane, which allows to use Normalized Device Coordinates (NDC) to bound all coordinates to a very small range(0-1).
In ~\tab{table_llff_rgb_uncertainty}, we report the performance of the different evaluated metrics averaged over all the scenes. As can be seen, S-NeRF achieves the second best performance in terms of uncertainty quantification  with quality metrics comparable to the rest of baselines. These results are consistent with the results reported in their original paper and demonstrate that the expressivity of S-NeRF on these simple scenes is not severely limited by the imposed strong assumptions over the radiance field distribution. Nonetheless, our CF-NeRF still achieves a significant improvement over most of the metrics, showing the advantages of modelling the radiance field distribution using the proposed latent variable modelling and CNFs.

\begin{table}[t]
\centering
\resizebox{\textwidth}{!}{%
\begin{tabular}{clccccccccc|cccccccc}
\hline
\multirow{3}{*}{\begin{tabular}[c]{@{}c@{}}Scene\\ Type\end{tabular}} & \multicolumn{1}{c}{\multirow{3}{*}{Methods}} &  & \multicolumn{3}{c}{Quality Metrics} &  & \multicolumn{3}{c}{Uncertainty Metrics} &  & \multirow{3}{*}{\begin{tabular}[c]{@{}c@{}}Scene\\ Type\end{tabular}} & \multicolumn{3}{c}{Quality Metrics} &  & \multicolumn{3}{c}{Uncertainty Metrics} \\ \cline{4-6} \cline{8-10} \cline{13-15} \cline{17-19} 
 & \multicolumn{1}{c}{} &  & \multirow{2}{*}{PSNR$\uparrow$} & \multirow{2}{*}{SSIM$\uparrow$} & \multirow{2}{*}{LPIPS$\downarrow$} &  & AUSE & AUSE & \multirow{2}{*}{NLL$\downarrow$} &  &  & \multirow{2}{*}{PSNR$\uparrow$} & \multirow{2}{*}{SSIM$\uparrow$} & \multirow{2}{*}{LPIPS$\downarrow$} &  & AUSE & AUSE & \multirow{2}{*}{NLL$\downarrow$} \\
 & \multicolumn{1}{c}{} &  &  &  &  &  & RMSE$\downarrow$ & MAE$\downarrow$ &  &  &  &  &  &  &  & RMSE$\downarrow$ & MAE$\downarrow$ &  \\ \hline
\multirow{5}{*}{Basket} & D.E.~\cite{deep_ensemble} &  & {\ul 25.93} & {\ul 0.87} & {\ul 0.17} &  & 0.206 & 0.161 & 1.26 &  & \multirow{5}{*}{Afria} & {\ul 23.14} & {\ul 0.78} & {\ul 0.32} &  & 0.385 & 0.304 & 2.19 \\
 & Drop.~\cite{mc_dropout} &  & 24.00 & 0.80 & 0.30 &  & 0.324 & 0.286 & 1.35 &  &  & 22.13 & 0.72 & 0.42 &  & 0.396 & 0.339 & 1.69 \\
 & NeRF-W~\cite{nerfw}&  & 19.81 & 0.73 & 0.31 &  & 0.333 & 0.161 & 2.23 &  &  & 21.14 & 0.71 & 0.42 &  & 0.121 & {\ul 0.089} & 1.41 \\
 & S-NeRF~\cite{snerf} &  & 23.56 & 0.80 & 0.23 &  & {\ul 0.098} & {\ul 0.089} & {\ul 0.26} &  &  & 20.83 & 0.73 & 0.35 &  & {\ul 0.209} & 0.161 & {\ul 0.56} \\
 & CF-NeRF &  & \textbf{26.39} & \textbf{0.89} & \textbf{0.11} &  & \textbf{0.039} & \textbf{0.018} & \textbf{-0.90} &  &  & \textbf{23.84} & \textbf{0.83} & \textbf{0.23} &  & \textbf{0.077} & \textbf{0.054} & \textbf{-0.25} \\ \hline
\multirow{5}{*}{Statue} & D.E.~\cite{deep_ensemble} &  & \textbf{24.57} & {\ul 0.85} & {\ul 0.21} &  & 0.307 & 0.214 & 1.53 &  & \multirow{5}{*}{Torch} & {\ul 21.49} & {\ul 0.73} & {\ul 0.43} &  & 0.153 & {\ul 0.101} & 1.80 \\
 & Drop.~\cite{mc_dropout} &  & 23.91 & 0.82 & 0.28 &  & 0.297 & 0.232 & {\ul 1.09} &  &  & 19.23 & 0.62 & 0.59 &  & 0.226 & 0.154 & 3.09 \\
 & NeRF-W~\cite{nerfw} &  & 19.89 & 0.73 & 0.41 &  & {\ul 0.099} & {\ul 0.071} & 3.03 &  &  & 15.59 & 0.56 & 0.69 &  & {\ul 0.132} & 0.131 & {\ul 1.52} \\
 & S-NeRF~\cite{snerf} &  & 13.24 & 0.55 & 0.59 &  & 0.475 & 0.714 & 4.56 &  &  & 13.12 & 0.33 & 1.02 &  & 0.321 & 0.454 & 2.29 \\
 & CF-NeRF &  & {\ul 24.54} & \textbf{0.87} & \textbf{0.16} &  & \textbf{0.040} & \textbf{0.019} & \textbf{-0.83} &  &  & \textbf{23.95} & \textbf{0.86} & \textbf{0.17} &  & \textbf{0.047} & \textbf{0.015} & \textbf{-0.86} \\ \hline
\end{tabular}%
}
\caption{Quality and uncertainty quantification metrics on rendered images over LF dataset. Best results are shown in bold with second best underlined. See text for more details.}
\label{tab:table_lf_rgb}
\end{table}
\begin{table}[t]
\centering
\resizebox{\textwidth}{!}{%
\begin{tabular}{clccccccccc|cccccccc}
\hline
 & \multicolumn{1}{c}{} &  & \multicolumn{3}{c}{Quality Metrics} &  & \multicolumn{3}{c}{Uncertainty Metrics} &  &  & \multicolumn{3}{c}{Quality Metrics} &  & \multicolumn{3}{c}{Uncertainty Metrics} \\ \cline{4-6} \cline{8-10} \cline{13-15} \cline{17-19} 
 & \multicolumn{1}{c}{} &  &  &  &  &  &  &  &  &  &  &  &  &  &  &  &  &  \\
\multirow{-3}{*}{\begin{tabular}[c]{@{}c@{}}Scene\\ Type\end{tabular}} & \multicolumn{1}{c}{\multirow{-3}{*}{Methods}} &  & \multirow{-2}{*}{RMSE$\downarrow$} & \multirow{-2}{*}{MAE$\downarrow$} & \multirow{-2}{*}{$\delta_3$ $\uparrow$} &  & \multirow{-2}{*}{\begin{tabular}[c]{@{}c@{}}AUSE \\ RMSE$\downarrow$ \end{tabular}} & \multirow{-2}{*}{\begin{tabular}[c]{@{}c@{}}AUSE \\ MAE$\downarrow$ \end{tabular}} & \multirow{-2}{*}{NLL$\downarrow$} &  & \multirow{-3}{*}{\begin{tabular}[c]{@{}c@{}}Scene\\ Type\end{tabular}} & \multirow{-2}{*}{RMSE$\downarrow$} & \multirow{-2}{*}{MAE$\downarrow$} & \multirow{-2}{*}{$\delta_3$ $\uparrow$} &  & \multirow{-2}{*}{\begin{tabular}[c]{@{}c@{}}AUSE \\ RMSE$\downarrow$ \end{tabular}} & \multirow{-2}{*}{\begin{tabular}[c]{@{}c@{}}AUSE \\ MAE$\downarrow$ \end{tabular}} & \multirow{-2}{*}{NLL$\downarrow$} \\ \hline
 & D.E.~\cite{deep_ensemble} &  & 0.221 & {\ul 0.132} & {\ul 0.66} &  & 0.480 & 0.232 & {\ul 7.88} &  &  & {\ul 0.085} & \textbf{0.039} & {\ul 0.90} &  & {\ul 0.218} & {\ul 0.110} & {\ul 3.61} \\
 & Drop.~\cite{mc_dropout} &  & 0.241 & 0.153 & 0.43 &  & {\ul 0.297} & {\ul 0.157} & 10.96 &  &  & 0.216 & 0.141 & 0.31 &  & 0.544 & 0.503 & 10.36 \\
 & NeRF-W~\cite{nerfw} &  & {\ul 0.214} & 0.179 & 0.41 &  & - & - & - &  &  & 0.127 & 0.079 & 0.77 &  & - & - & - \\
 & S-NeRF~\cite{snerf} &  & 0.417 & 0.401 & 0.23 &  & 0.305 & 0.312 & 8.75 &  &  & 0.239 & 0.155 & 0.36 &  & 0.234 & 0.252 & 6.55 \\
\multirow{-5}{*}{Basket} & CF-NeRF &  & \textbf{0.166} & \textbf{0.099} & \textbf{0.80} &  & \textbf{0.101} & \textbf{0.052} & \textbf{6.76} &  & \multirow{-5}{*}{Africa} & {\color[HTML]{000000} \textbf{0.074}} & \textbf{0.039} & \textbf{0.93} &  & \textbf{0.105} & \textbf{0.090} & \textbf{2.05} \\ \hline
 & D.E.~\cite{deep_ensemble} &  & {\ul 0.162} & {\ul 0.118} & {\color[HTML]{333333} \textbf{0.73}} &  & {\ul 0.115} & {\ul 0.056} & {\ul 7.78} &  &  & {\ul 0.132} & {\ul 0.071} & {\ul 0.71} &  & {\ul 0.226} & {\ul 0.131} & {\ul 5.76} \\
 & Drop.~\cite{mc_dropout} &  & 0.197 & 0.128 & 0.59 &  & 0.164 & 0.109 & 10.89 &  &  & 0.263 & 0.173 & 0.06 &  & 0.982 & 0.809 & 10.84 \\
 & NeRF-W~\cite{nerfw} &  & 0.276 & 0.218 & 0.55 &  & - & - & - &  &  & 0.271 & 0.241 & 0.17 &  & - & - & - \\
 & S-NeRF~\cite{snerf} &  & 0.751 & 0.709 & 0.26 &  & 0.353 & 0.386 & 11.51 &  &  & 0.274 & 0.236 & 0.14 &  & 0.770 & 1.013 & 8.75 \\
\multirow{-5}{*}{Statue} & CF-NeRF &  & \textbf{0.122} & \textbf{0.098} & {\color[HTML]{333333} \textbf{0.73}} &  & \textbf{0.069} & \textbf{0.053} & \textbf{7.38} &  & \multirow{-5}{*}{Torch} & \textbf{0.110} & \textbf{0.061} & \textbf{0.78} &  & \textbf{0.164} & \textbf{0.089} & \textbf{4.15} \\ \hline
\end{tabular}}
\caption{Quality and uncertainty quantification metrics on depth estimation over the LF dataset.  All the results are computed from disparity values which are reciprocal to depth. Best results are shown in bold with second best underlined. See text for more details.}
\label{tab:table_lf_depth}
\end{table}

\subsection{Results over LF dataset}

To further test the performance of different approaches on larger scenes with more complex appearances and geometries, we evaluate the different methods on the LF dataset. All images in these scenes have a large depth range and are randomly captured towards the target. As a consequence, coordinates cannot be normalized with NDC.

\mypar{Quantitative results} Results over rendered RGB images and estimated depth-maps for all the scenes and evaluated metrics are shown in \tab{table_lf_rgb} and \tab{table_lf_depth}, respectively.  As can be seen, S-NeRF obtains poor performance in terms of quality  and uncertainty estimation. This clearly shows that the imposed assumptions over the radiance fields distribution are sub-optimal in complex scenes and lead to inaccurate model predictions and unreliable uncertainty estimations. In contrast, our CF-NeRF outperforms all the previous approaches by large margins across all the scenes and metrics both on image rendering and depth-maps estimation tasks. The better results obtained by our method can be explained because the use of latent variable modelling and conditional normalizing flows allows to learn arbitrarily complex radiance field distributions, which are necessary to model scenes with complicated appearance and geometry. This partly explains that our method achieves better performance, particularly an average 69$\%$ improvement compared to the second best method Deep-Ensembles on LPIPS, which measures the perceptual similarity at the image level.  On the other hand, the performance of Deep-Ensembles and MC-Dropout is largely limited by the number of the trained models or dropout samples, which cannot be increased infinitely in practical cases. In contrast, the explicitly encoding of the radiance field distribution into a single probabilistic model with CF-NeRF allows to use a higher-number of samples in a more efficient manner. 

\begin{figure}[t]
\centering
  \includegraphics[width=1.0\textwidth]{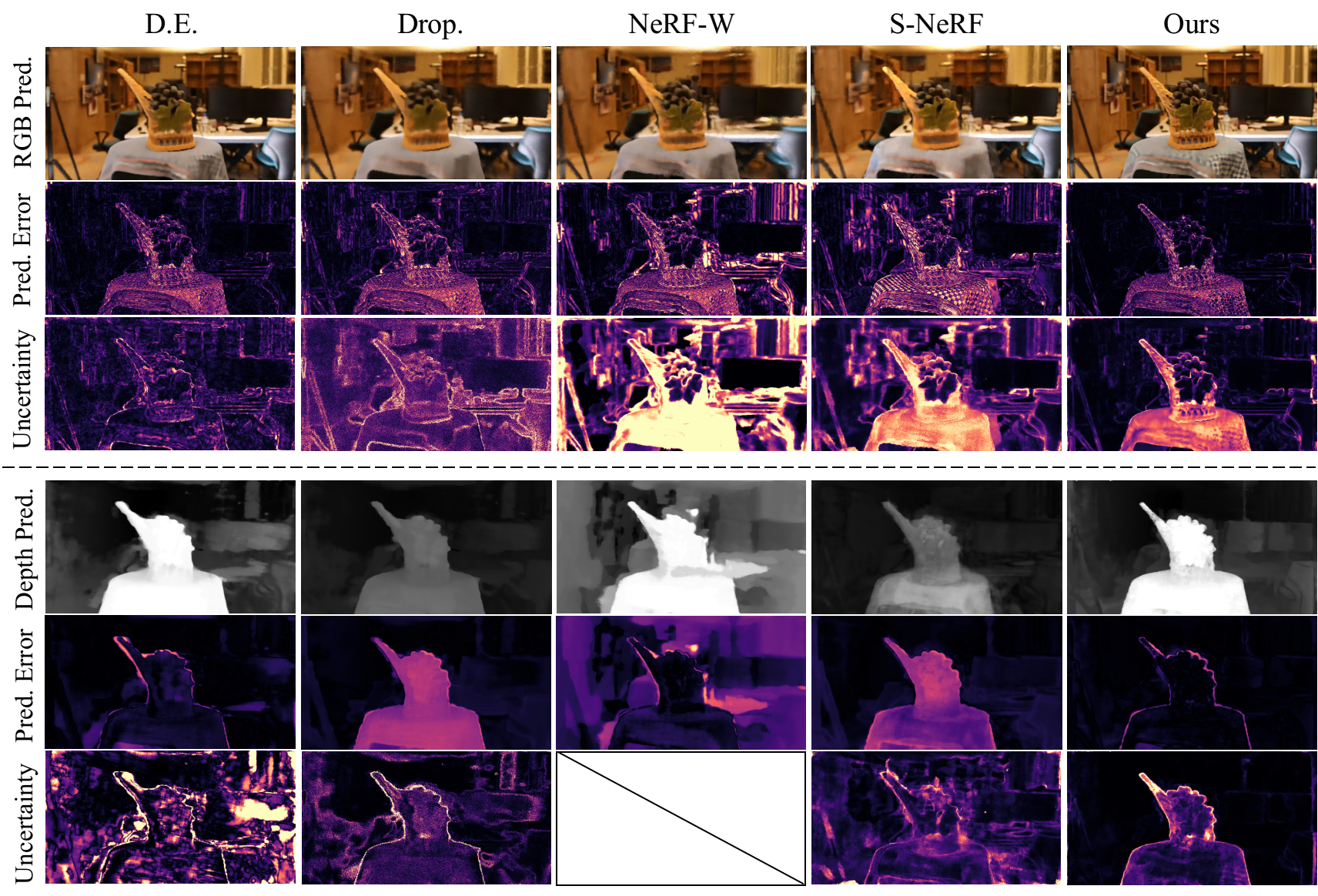}
  \vspace{-3mm}
  \caption{Qualitative comparison between our CF-NeRF and other baselines over generated images(Top) and depth-maps(Bottom). 
  The model prediction, its computed error with the Ground-Truth and its associated uncertainty estimation are shown respectively in each row.  Larger values are shown in yellow while purple and black indicate smaller values of predicted error and uncertainty. NeRF-W is not able to generate multiple depth samples and hence cannot produce uncertainty for depth-maps. Among all these approaches, our CF-NeRF is the only approach that is able to generate both accurately rendered views and depth estimations and visually intuitive uncertainty maps which are highly correlated with the true prediction error. Refer to the supplementary materials for more visualized results.}
  \label{fig:comparison}
\end{figure}

\mypar{Qualitative results} In order to give more insights into the advantages of our proposed CF-NeRF, \fig{comparison} shows qualitative results obtained by the compared methods on an example testing view of the LF dataset. By looking at the estimated uncertainty for RGB images, we can observe that NeRF-W obtains a higher prediction error which, additionally, is not correlated with the uncertainty estimates. On the other hand, Deep Ensembles and MC-Dropout render images with a lower predictive error compared to NeRF-W. As can be observed, however, their estimated uncertainties are noticeably not correlated with their errors. Interestingly, S-NeRF seems to obtain uncertainty estimates where the values are more correlated with the predictive error. However, the quality of the rendered image is poor, obviously shown in the background. In contrast, our CF-NeRF is the only method able to render both high-quality images and uncertainty estimates that better correlate with the prediction error. 

By looking at the results for generated depth-maps, we can see that our method also obtains the most accurate depth estimates, followed by Deep Ensembles. Nonetheless, the latter generates depth maps with obvious errors in the background. Regarding the uncertainty estimation, we can observe a high correlation between the predicted error and the estimated uncertainty maps generated by our CF-NeRF. This contrasts with the rest methods which estimate low confidence values in the background of the scene inconsistent with their low prediction errors on this specific area. In conclusion, our results demonstrate the ability of our proposed method to provide both reliable uncertainty estimates and accurate model predictions.

\section{Conclusion}

We have presented CF-NeRF, a novel probabilistic model to address the problem of uncertainty quantification in 3D modelling using Neural Radiance Fields. In contrast to previous works employing models with limited expressivity, our method couples Latent Variable Modelling and Conditional Normalizing Flows in order to learn complex radiance fields distributions in a flexible and fully data-driven manner. In our experimental results, we show that this strategy allows CF-NeRF to obtain significantly better results than state-of-the-art approaches, particularly on scenes with complicated appearance and geometry. Finally, it is also worth mentioning that our proposed framework can be easily combined with other NeRF-based architectures in order to provide them with uncertainty quantification capabilities.


%
\clearpage
\bibliographystyle{splncs04}
\bibliography{egbib}

\clearpage
\begin{subappendices}
\onecolumn
\clearpage
\appendix

\section*{Supplementary Materials}

In this Supplementary Materials, we firstly give more additional details about our CF-NeRF implementation (\sect{implementation}), then we describe a set of ablation studies to give more insights about the performance of our model (\sect{ablations}) and, finally, we provide a set of additional qualitative results (\sect{more_results}). 

\section{Additional Implementation Details}
\label{sec:implementation}

\subsubsection{Training details}
As mentioned in \sect{implementations}, we use the same MLP-based architecture used in original NeRF \cite{nerf} as a backbone network for our CF-NeRF and the rest baselines. In particular, we use 512 hidden units for all layers. For CF-NeRF, each sample from the latent prior distribution is shared for different spatial-location and viewing-direction inputs in each batch during training. To avoid overfitting with the sparse number of training views used in our experiments, we employ an additional depth loss based on \cite{nerfds} during optimization. This loss is weighted with a value of $1e-2$ for our method and the rest baselines. Additionally, we set a value of 0.01 as the weight for the Entropy term in \Eq{kl_fnerf}. 

\subsubsection{Conditional Normalizing Flows}
As for invertible transformation functions in our Conditional Normalizing Flow(CNF), we use the Sylvester Flows \cite{sylvester} defined as:
\begin{equation}
    \textbf{z}_k = \textbf{z}_{k-1} + \textbf{A} h(\textbf{B} \textbf{z}_{k-1} + b),
\end{equation}
where $\textbf{A}$, $\textbf{B}$ and $\textbf{b}$ are flow parameters of each transformation function. Additionally, $h$ is an hyperbolic tangent activate function. 
These flow parameters are conditional functions of the 5D location-direction pairs, while the samples from the latent distributions are transformed to radiance and density by sequentially using these transformation functions $f_{1:K}$ described in \sect{conditional_flow}.
In our CF-NeRF, we use four flows for the radiance and density CNFs (see \fig{conditional_nf}) with the dimensions of the conditional feature into each flow set to 64. 

\subsubsection{Metrics}
As a metric used to assess the quality of the depth prediction, we use the $\delta$-threshold \cite{delta_threshold}. This metric is defined as follows:
\begin{equation}
    \% \text{    of    } y_i \text{    s.t.    } max(\frac{y_i^*}{y_i}, \frac{y_i}{y_i^*}) = \delta_{k=1,2,3} < \tau^{k=1,2,3},
\end{equation}
where we set the threshold $\tau=1.25$ as done in previous works \cite{delta_threshold}. Note that we only report $\delta_3$ due to space limitations in the main paper.

\section{Ablations}
\label{sec:ablations}

In order to give insights into some design decisions of our proposed CF-NeRF, we provide results for two ablation experiments. Concretely, we conduct experiments over the LF dataset. Results are shown in \tab{ablations}. In the following, we describe each of the experiments in more detail.

\mypar{Entropy term} 
We remove the entropy term in \Eq{kl_fnerf} and train our CF-NeRF only using the NLL as the training loss. On both generated RGB images and depth-maps, we achieve better performance by using the Entropy term as well across all metrics, including the prediction error and its associated uncertainty. This is consistent with what we have discussed in \sect{optimization} that, maximizing the Entropy term intuitively prevents the optimized distribution to degenerate into a deterministic function where all the probability is assigned into a single radiance field $\mathcal{F}$, thus losing the ability to quantify correct uncertainty. 

\mypar{Single Flow}
As discussed in \sect{cf_nerf}, our CF-NeRF uses two conditional normalizing flows(CNF) for modelling the distribution of radiance and density. However, a more efficient strategy could be to jointly model their distributions using a single flow in order to take into account the possible dependence between them. As we can see in \tab{ablations}, this variant obtains worse performance compared to our CF-NeRF with two CNFs in terms of prediction quality and uncertainty estimation. This drop in performance is especially high in the case of depth-map estimation. This can be explained because using a single CNF for radiance and density distribution contradicts the fact that the volume density must be independent of the emitted radiance to obtain optimal results, as was previously discussed in \cite{nerf,nerf++}.

\begin{table}[t]
\centering
\resizebox{\textwidth}{!}{%
\begin{tabular}{llccccccccc}
\hline
 & \multirow{3}{*}{Methods} &  & \multicolumn{3}{c}{Quality Metrics} &  & \multicolumn{3}{c}{Uncertainty Metrics} &  \\ \cline{4-6} \cline{8-10}
 &  &  & \multirow{2}{*}{PSNR$\uparrow$} & \multirow{2}{*}{SSIM$\uparrow$} & \multirow{2}{*}{LPIPS$\downarrow$} &  & \multirow{2}{*}{AUSE RMSE$\downarrow$} & \multirow{2}{*}{AUSE MAE$\downarrow$} & \multirow{2}{*}{NLL$\downarrow$} &  \\
 &  &  &  &  &  &  &  &  &  &  \\ \hline
\multirow{3}{*}{RGB images} & CF-NeRF  w/o Entropy &  & 23.40 & 0.81 & 0.258 &  & 0.068 & 0.048 & -0.448 &  \\
 & CF-NeRF  w/ Single Flow &  & 23.82 & 0.83 & 0.228 &  & 0.081 & 0.039 & -0.578 &  \\
 & CF-NeRF &  & \textbf{24.78} & \textbf{0.86} & \textbf{0.168} &  & \textbf{0.051} & \textbf{0.026} & \textbf{-0.710} &  \\ \hline
 & \multirow{2}{*}{} & \multicolumn{1}{l}{} & \multirow{2}{*}{RMSE$\downarrow$} & \multirow{2}{*}{MAE$\downarrow$} & \multirow{2}{*}{$\delta_3$ $\uparrow$} & \multicolumn{1}{l}{} & \multirow{2}{*}{AUSE RMSE$\downarrow$} & \multirow{2}{*}{AUSE MAE$\downarrow$} & \multirow{2}{*}{NLL$\downarrow$} & \multicolumn{1}{l}{} \\
 &  & \multicolumn{1}{l}{} &  &  &  & \multicolumn{1}{l}{} &  &  &  & \multicolumn{1}{l}{} \\ \hline
\multirow{3}{*}{Depth} & CF-NeRF  w/o Entropy & \multicolumn{1}{l}{} & 0.121 & 0.078 & 0.76 & \multicolumn{1}{l}{} & 0.224 & 0.143 & 7.88 & \multicolumn{1}{l}{} \\
 & CF-NeRF  w/ Single Flow & \multicolumn{1}{l}{} & 0.170 & 0.111 & 0.64 & \multicolumn{1}{l}{} & 0.229 & 0.138 & 8.16 & \multicolumn{1}{l}{} \\
 & CF-NeRF & \multicolumn{1}{l}{} & \textbf{0.118} & \textbf{0.074} & \textbf{0.81} & \multicolumn{1}{l}{} & \textbf{0.110} & \textbf{0.071} & \textbf{5.09} & \multicolumn{1}{l}{} \\ \hline
\end{tabular}%
}
\caption{Results of our ablation studies: Quality and uncertainty quantification metrics on rendered images and depth-maps over LF dataset. Best results are shown in bold. See text for more details.}
\label{tab:ablations}
\end{table}

\section{More Results}
\label{sec:more_results}

\subsubsection{Interpolation videos}
An intuitive advantage of the explicit distribution modelling over the radiance fields in our CF-NeRF is that, we can conveniently analyze the learned radiance fields by interpolating in the latent space ~\cite{interpolations1,interpolations2}. 
The shared latent variable allows to model the joint distribution of all the radiance-density pairs in the scene in contrast to S-NeRF and hence could avoid the noisy results as discussed in \sect{intro} and illustrated in \fig{fig2}. 
More formally, we define the interpolation value as,
\begin{equation}
    f_{L}(\textbf{z}_1,\textbf{z}_2,\lambda) = \lambda \textbf{z}_1 + (1-\lambda) \textbf{z}_2
\end{equation}
where $\textbf{z}_1$ and $\textbf{z}_2$ are two random samples from the latent distribution with $\lambda \in [0,1]$. Then the density and radiance can be obtained through our proposed CNF, following our inference process as discussed in \sect{inference} to render novel views and depth.
To see the dynamic interpolation results we provide a video attached to this supplementary material. 
By looking at different frames in the dynamic interpolated results, S-NeRF tends to generate noisy image and depth predictions with random and incoherent changes between adjacent frames obtained using two adjacent interpolation values. In contrast, our CF-NeRF can generate more coherent and smoothly changing frames, both on rendered RGB images and estimated depth-maps. This clearly demonstrates the advantages of our proposed Latent Variable Modelling for CF-NeRF in order to efficiently model the joint distribution over all the possible radiance and density pairs in the scene.

\begin{figure*}[t]
  \centering
  \includegraphics[width=1.0\textwidth]{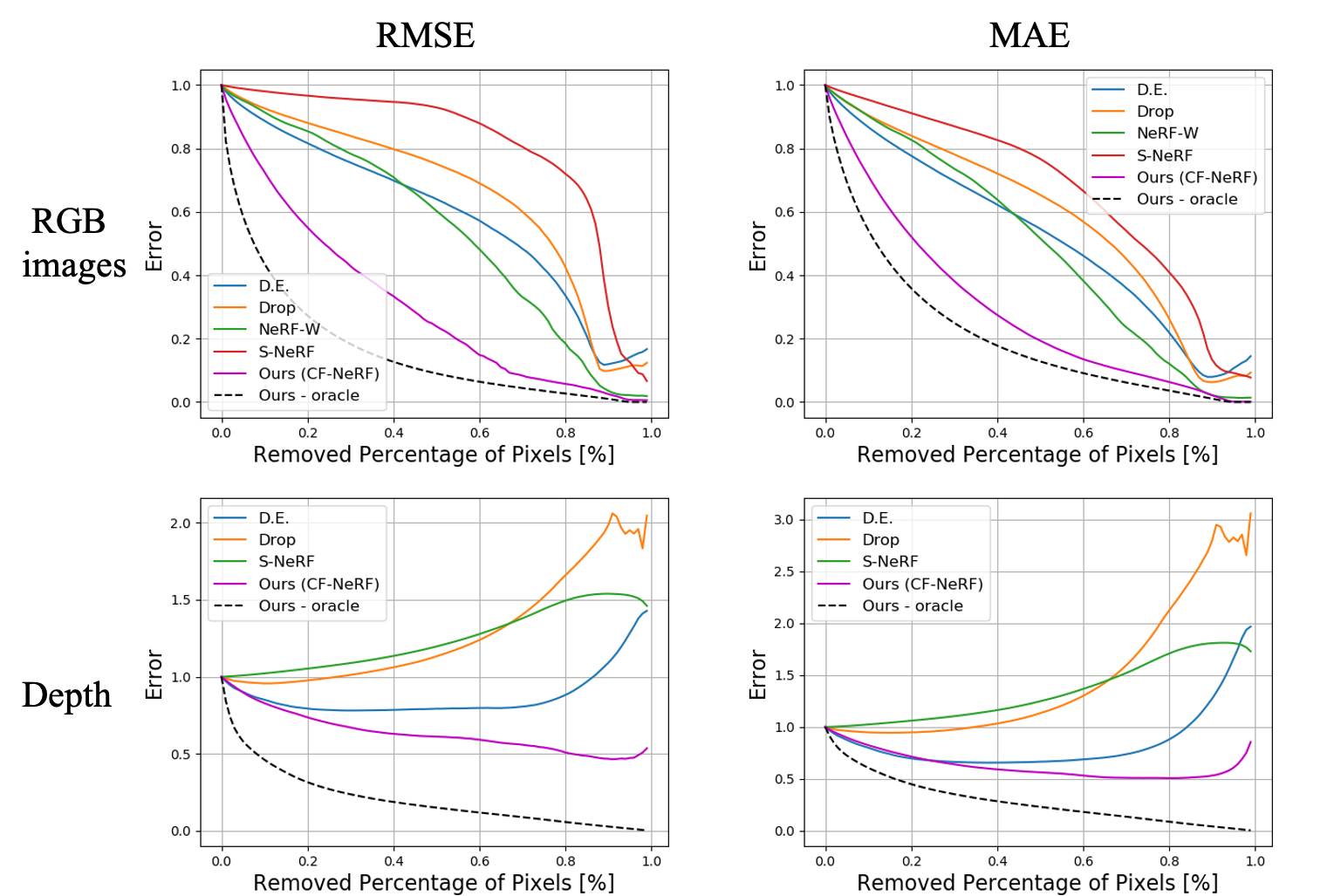}
  \caption{Sparsification curves obtained by different methods of estimating uncertainty associated with rendered RGB images and estimated depth.} 
  \label{fig:ause_rgb}
\end{figure*}

\subsubsection{Sparsification plots}
\fig{ause_rgb} shows the additional related sparsification curves on the synthetic novel views and estimated depth averagely over the LF dataset. Note that NeRF-W is not able to estimate uncertainty on depth as analyzed in \sect{related_work} and hence cannot generate the sparsification curve on depth. When evaluated over all pixels, all methods perform similarly. As we remove the pixels with high uncertainty from 1$\%$ to 100$\%$, our method always obtains the lowest value and fits closest with the oracle curve. This demonstrates that our estimated uncertainty correlates significantly better with the prediction error than the others.

\subsubsection{More qualitative results}
\begin{figure*}[t]
  \centering
  \includegraphics[width=1.0\textwidth]{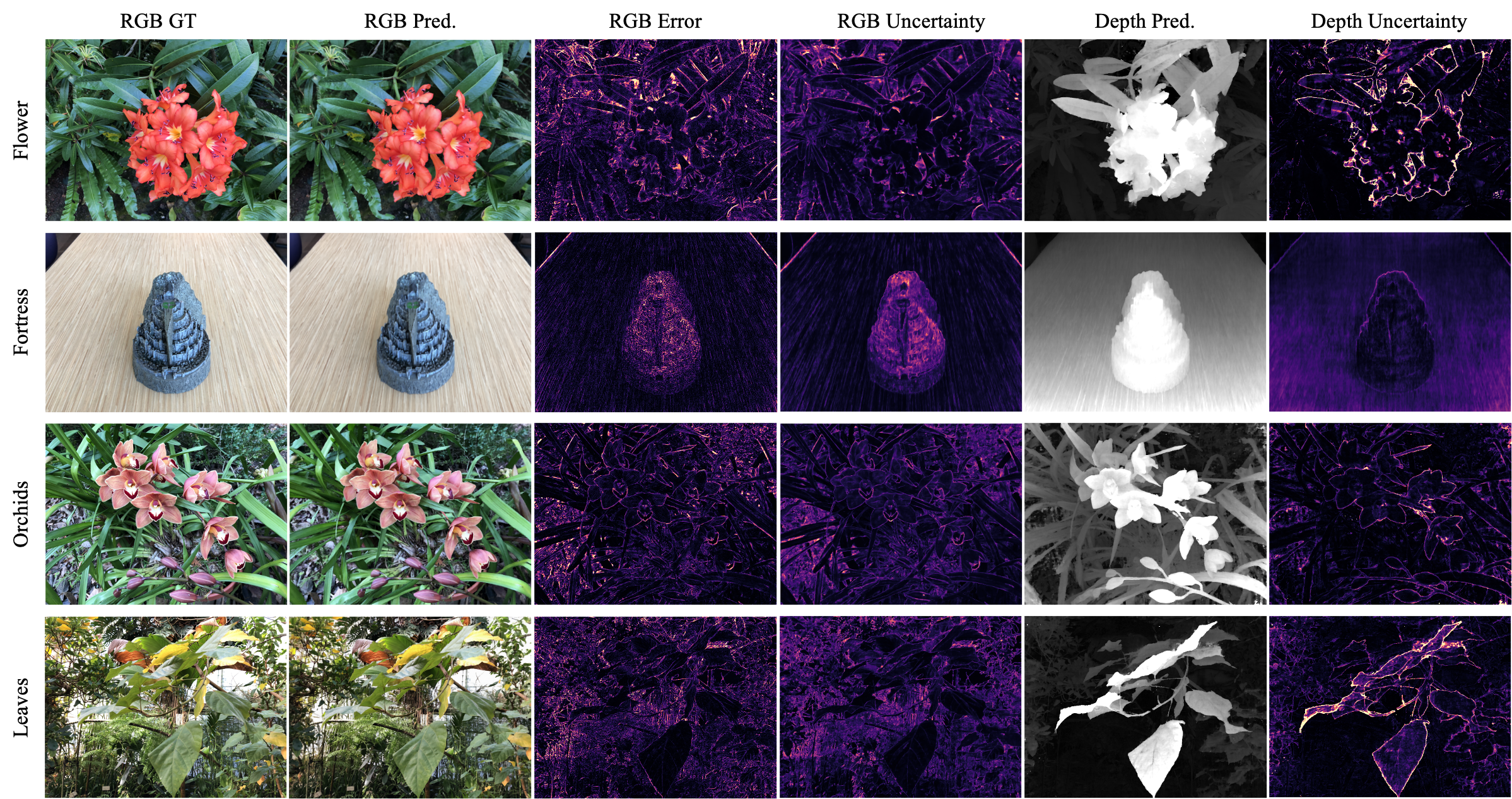}
  \caption{More qualitative results obtained by our CF-NeRF over LLFF dataset.} 
  \label{fig:supplem_llff}
\end{figure*}

\begin{figure*}[t]
  \centering
  \includegraphics[width=1.0\textwidth]{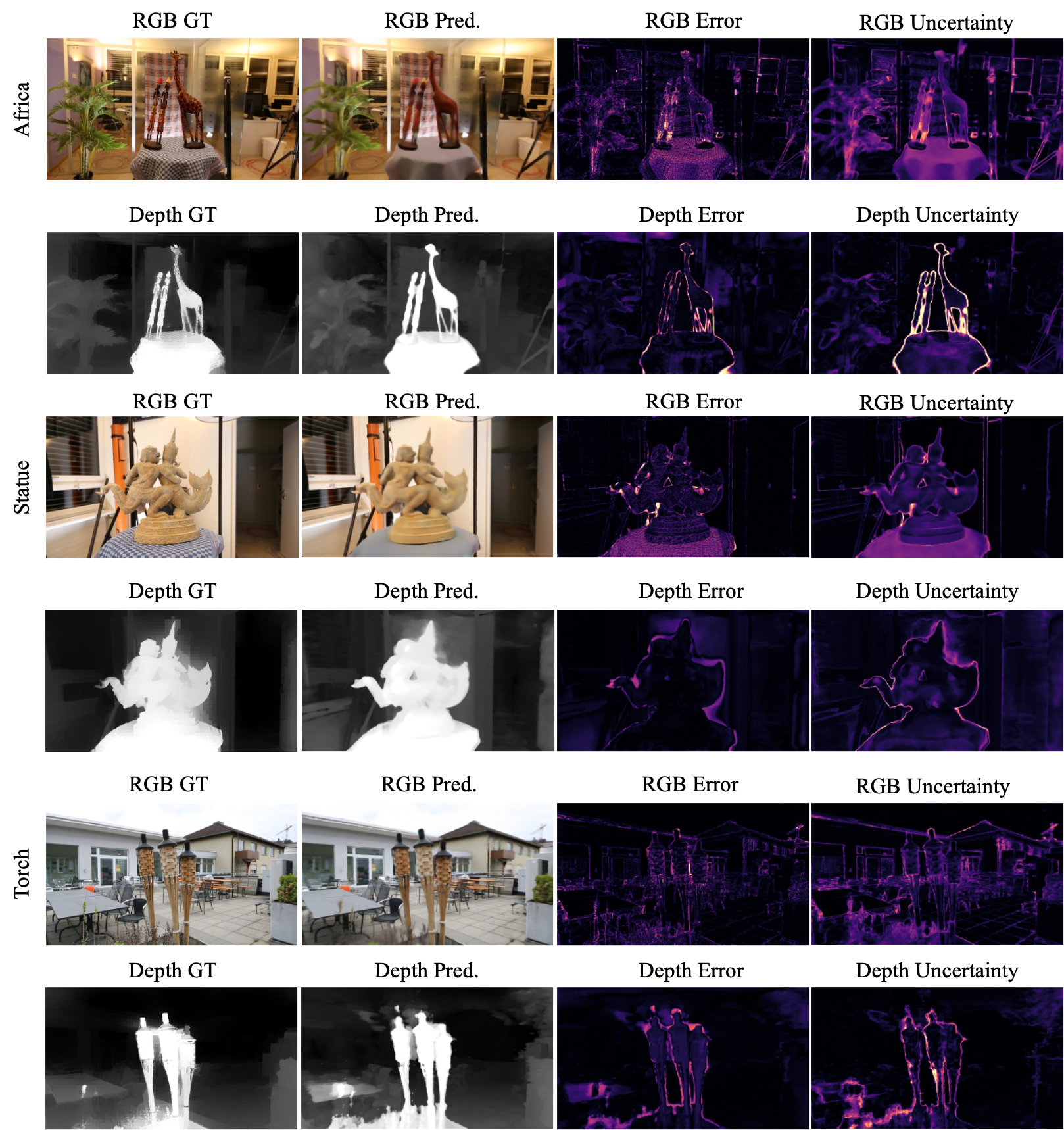}
  \caption{More results obtained by our CF-NeRF over LF dataset.} 
  \label{fig:supplem_lf}
\end{figure*}

\fig{supplem_llff} shows more qualitative results obtained by our CF-NeRF for the scenes in the simple LLFF dataset.
Moreover, \fig{supplem_lf} shows additional qualitative results obtained by our CF-NeRF across other scenes in the LF dataset: \textit{Africa, Statue, Torch}.
For each scene, we show not only the predicted RGB views and the estimated depth-maps, but also their associated uncertainty estimations.

\end{subappendices}

\end{document}